\documentclass[10pt,twocolumn]{ICCAS2024}
 
\usepackage{kotex}
\usepackage{cite}
\usepackage{diagbox}
\usepackage{mdwmath}
\usepackage{algorithm}
\usepackage{algpseudocode}
\usepackage{hyperref}
\usepackage[utf8]{inputenc}

\begin{document}

\title{ROFT-VINS: Robust Feature Tracking-based\\ Visual-Inertial State Estimation for Harsh Environment}

\author{Sanghyun Park$^{1}$ and Soohee Han$^{2*}$ }

\affils{ ${}^{1}$Department of Convergence IT Engineering, POSTECH, \\
Pohang, 37673, Korea (pash0302@postech.ac.kr) \\
${}^{2}$Department of Electrical Engineering and of Convergence IT Engineering, POSTECH, \\
Pohang, 37673, Korea (sooheehan@postech.ac.kr) {\small${}^{*}$ Corresponding author}}


\abstract{
    SLAM (Simultaneous Localization and Mapping) and Odometry are important systems for estimating the position of mobile devices, such as robots and cars, utilizing one or more sensors. Particularly in camera-based SLAM or Odometry, effectively tracking visual features is important as it significantly impacts system performance. In this paper, we propose a method that leverages deep learning to robustly track visual features in monocular camera images. This method operates reliably even in textureless environments and situations with rapid lighting changes. Additionally, we evaluate the performance of our proposed method by integrating it into VINS-Fusion (Monocular-Inertial), a commonly used Visual-Inertial Odometry (VIO) system.
}


\maketitle


\section{Introduction}

SLAM and odometry are important systems that enable autonomous navigation\cite{nav} in mobile robots and self-driving vehicles. These rely on accurate estimation of the device's position and orientation using various sensor inputs such as cameras, LiDAR, and Inertial Measurement Units (IMU). Among these, Visual-Inertial Odometry (VIO) integrates information from cameras and IMU, making it robust against challenges such as textureless environments and motion blur. In the case of monocular systems, VIO also allows for the estimation of scale information\cite{ORBSLAM3_TRO}.

In visual-based odometry, the ability to reliably track features across consecutive frames is important. Traditional methods often rely on feature descriptors such as SIFT\cite{SIFT}, SURF\cite{SURF}, and ORB\cite{ORB}, which have shown robustness in general environments. However, these methods can encounter challenges in environments with textureless or rapidly changing lighting conditions. These drive the investigation of sophisticated methods to enhance the reliability and precision of feature tracking.

Deep learning has provided powerful tools and driven innovation in many areas of computer vision, such as object detection, segmentation, and optical flow estimation. Optical flow, the pattern of apparent motion of objects in a visual scene, is particularly relevant for visual SLAM and odometry. Recent advancements in deep learning-based optical flow estimation, such as the RAFT (Recurrent All-Pairs Field Transforms) network\cite{RAFT}, have shown remarkable performance improvements over traditional methods. RAFT's iterative refinement process\cite{RAFT} and all-pairs field transforms enable highly accurate and dense optical flow estimation, performing well even in challenging scenarios.

VINS-Mono\cite{VINS} is a widely used VIO system in both academia and industry due to its robust and accurate performance. VINS-Mono\cite{VINS} leverages visual and inertial measurements to provide pose estimation. However, like many VIO systems, its performance heavily depends on the quality of visual feature tracking. Therefore, improving the feature tracker in VINS-Mono\cite{VINS} can enhance the overall system performance.

The VINS-Fusion system experimented on in this paper is an extended version of VINS-Mono\cite{VINS}. VINS-Fusion not only expands the original VINS-Mono\cite{VINS}, which was limited to monocular inertial systems, to stereo inertial systems but also implements a multi-threading. Since the monocular inertial system in VINS-Fusion is nearly identical to that of VINS-Mono\cite{VINS}, VINS-Fusion as mentioned in this paper refers to the VINS-Mono\cite{VINS} system.

In this paper, we propose integrating the RAFT\cite{RAFT}-based optical flow method into the feature tracker of VINS-Mono\cite{VINS}. By leveraging RAFT's\cite{RAFT} superior optical flow estimation capabilities, our goal is to enhance the robustness and accuracy of VINS-Mono\cite{VINS} in environments with sparse textures or dynamic lighting conditions. Additionally, through outlier rejection, we provide a more accurate feature tracker performance.

The structure of this paper is as follows. In section 2, we review related work on VIO and deep learning-based optical flow. In Section 3, we detail the method of integrating RAFT\cite{RAFT} into the feature tracking pipeline of VINS-Mono\cite{VINS}. In section 4, we present our experimental setup and results, demonstrating the performance improvements achieved by the proposed approach. Finally, we make a conclusion in Section 5.

By utilizing deep learning-based visual feature tracking, this paper contributes to the fields of autonomous driving and robotics, providing a foundation for more robust and reliable VIO systems. These contributions enable autonomous vehicles and robots to operate more stably in diverse environments, ultimately playing a crucial role in developing safer and more efficient autonomous driving solutions.

\section{Related work}

In this section, We review the VIO and deep learning-based optical flow.

\subsection{Visual-Inertial Odometry}

SLAM and odometry are both methods used in robotics to estimate the position of mobile devices. However, they differ in their purpose and methodology. Odometry is tracking the position of a robot or vehicle as it moves, primarily using encoders, IMUs, and other sensors. This method estimates the current position by accumulating the traveled distance from an initial position. However, odometry can become less accurate over time due to accumulated drift. In contrast, SLAM allows a robot to simultaneously estimate its position and map an unknown environment. SLAM updates both the position and the map by fusing various sensor data such as cameras, LiDAR, and IMU. During this process, the robot recognizes landmarks in the environment and recalculates its position based on them, thereby reducing errors.

VIO is a method that fuses data from cameras and IMU to estimate the position and orientation of a device. VIO has an important function in various applications like robotics, autonomous driving, and AR(Augmented Reality)\cite{PWCNet}.

VINS-Mono\cite{VINS} is versatile and robust monocular inertial SLAM system. It includes loop closing with DBoW2\cite{BOW} and 4 DoF pose-graph optimization. Also, it has feature tracker using Lucas-Kanade\cite{LK} method based optical flow.

ORB-SLAM3\cite{ORBSLAM3_TRO} is one of the latest versions of visual-inertial SLAM systems, performing real-time localization and mapping in 3D environments. It supports monocular, stereo, and RGB-D cameras, and has real-time operation, robust localization, loop closing, and relocalization capabilities. It can adapt to environmental changes over time and efficiently operate in large or complex environments through its multi-map system. Utilizing the ORB\cite{ORB} feature extractor and keyframe-based tracking, it estimates the relative motion of the camera with visual-inertial odometry. ORB-SLAM3\cite{ORBSLAM3_TRO} is a visual-inertial SLAM system that operates stably and accurately in various environments, providing real-time performance.

\subsection{Deep Learning based Optical Flow}

Optical flow estimates pixel motion between consecutive image frames to understand the pattern of apparent motion. Traditional optical flow algorithms are primarily based on mathematical models, but advancements in deep learning have led to the emergence of more powerful and accurate methods.

RAFT\cite{RAFT} is a deep learning model for optical flow estimation that accurately tracks the movement of objects in image sequences. RAFT\cite{RAFT} estimates motion vectors by iteratively computing correlations between all pairs of pixels, providing high-precision optical flow estimation and efficient computation. This enables optimal optical flow estimation through end-to-end learning and demonstrates excellent performance across various datasets. Its main components include a grid feature extractor for extracting image features, a cost volume generation for calculating pixel pair correlations between images, and a recurrent update module that iteratively updates the optical flow using a recurrent network. RAFT\cite{RAFT} can be utilized in various fields such as computer vision, autonomous driving, and augmented reality, excelling particularly in applications like video analysis, object tracking, and motion compensation.

Tracking Everything Everywhere All at Once\cite{omnimotion}, called OmniMotion, is an advanced video tracking method designed to estimate dense, long-range motion across entire video sequences. OmniMotion\cite{omnimotion} addresses the limitations of traditional optical flow and particle video tracking algorithms, which often struggle with occlusions and maintaining global consistency over extended sequences. By using a quasi-3D canonical volume representation, OmniMotion\cite{omnimotion} ensures globally consistent tracking of every pixel throughout a video, even when objects are occluded. This is achieved through bijective mappings between local frames and the canonical volume, allowing accurate tracking of motion without needing explicit 3D reconstruction of the scene. The system is built on a neural network architecture that includes six affine coupling layers and uses a coordinate-based network for mapping 3D coordinates to density and color. This enables the algorithm to retain information about the relative depth of scene points, which helps in maintaining accurate tracking through occlusions.

\section{Method}

\begin{figure}[thb]
\begin{center}
\label{architecture}
\includegraphics[width=7.8cm]{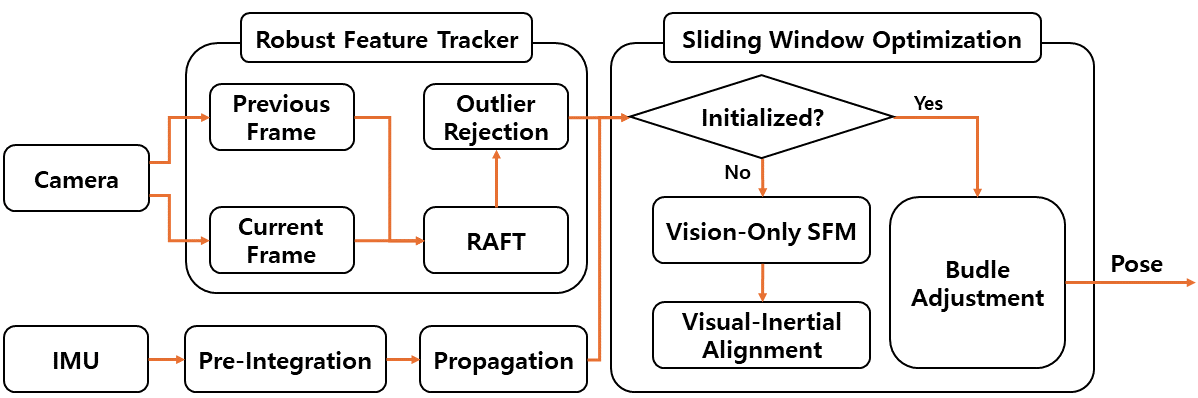}
\caption{The architecture of our odometry system.}
\end{center}
\end{figure}

\begin{figure}[thb]
\begin{center}
\label{architecture}
\includegraphics[width=7.8cm]{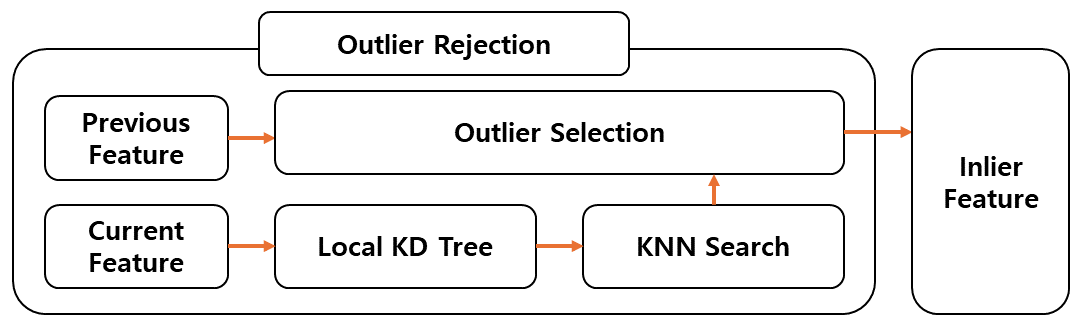}
\caption{The pipeline of our outlier rejection module.}
\end{center}
\end{figure}

Our proposed method is shown in Fig. 1. We use the odometry estimation part of VINS-Mono\cite{VINS}. In this setup, we have integrated RAFT\cite{RAFT} into the feature tracker system. Additionally, in the outlier rejection part, features that exhibit different movements from their surrounding features are identified as outliers and subsequently removed.

\subsection{Robust Feature Tracker}

In VINS-Mono\cite{VINS}, the KLT sparse optical flow algorithm\cite{LK} based on the Lucas-Kanade method is used for optical flow. Additionally, new features are detected\cite{GoodFeature} to maintain a sufficient number of features in each image\cite{VINS}. Furthermore, when detecting features, they are selected to ensure a minimum distance between them\cite{VINS}.

Our approach uses RAFT\cite{RAFT} instead of \cite{LK} for optical flow estimation. This method is robust against aggressive motion as well as unstable image changes caused by sudden variations in lighting. However, because RAFT\cite{RAFT} provides results for the all pixel rather than specific pixels, post-processing is required. First, we identify the pixels corresponding to the features, then extract the motion information for these pixels from the RAFT\cite{RAFT} inference results.

The robust feature tracker process is presented in Algorithm 1. By applying the RAFT\cite{RAFT} inference image to the previous frame's features $f_{t-1}$, we obtain the tracked features $f_t$ in the current frame. Outliers that fall outside the image boundaries are removed, and new features are extracted using \cite{GoodFeature} to maintain the number of features, equivalent to the number of removed features. During this process, a mask is generated for the existing features to ensure that the distance between features is above a certain threshold. Finally, for use in the state estimator, we accumulate the results of $f_t$, the undistorted results of $f_t$, and the velocity between $f_t$ and $f_{t-1}$.

The outlier rejection pipeline is shown in Fig. 2. First, 2D KD tree is constructed using the tracked features. Then, for each feature, the KD tree is used to search for the nearest $n$ features within a certain distance. If a feature exhibits a different movement compared to its surrounding features, based on the tracked information, it is identified as an outlier and removed.

\begin{algorithm}
\caption{Robust Feature Tracker}
\begin{algorithmic}[1]
\State \textbf{Input:} Image from current frame $I_t$
\Procedure{Track Feature}{$I_t$}
    \State Image from previous frame $I_{t-1}$
    \State Tracked feature from previous frame $f_{t-1}$
    \State $f_t \leftarrow$ GetTrackResult($I_{t-1},I_t,f_{t-1}$)
    \State $f_t \leftarrow$ OutlierRejection($f_t$)
    \State Add feature for reduced feature num:
    \State \indent $f_t \leftarrow$ $f_t +$ GoodFeatureToTrack($I_t$)
    \State Get feature information $F_t$ for state estimation:
    \State \indent $F_t \leftarrow$ GetFeatureInfo($f_t,f_{t-1},I_t,I_{t-1}$)
    \State $I_{t-1} \leftarrow I_t$
    \State $f_{t-1} \leftarrow f_t$
    \State \Return $F_t$
\EndProcedure
\Function {}{}GetTrackResult($I_{t-1},I_t,f_{t-1}$)
    \State Get RAFT result image $R$:
    \State \indent $R \leftarrow$ RAFT$(I_{t-1},I_t,f_{t-1})$
    \State Calculate tracked feature $f_t$:
    \State \indent $f_t \leftarrow R(f_{t-1})$
    \State \Return $f_t$
\EndFunction
\Function {}{}GetFeatureInfo($f_t,f_{t-1},I_t,I_{t-1}$)
    \State $F_t \leftarrow$ $f_t$
    \State $F_t \leftarrow F_t +$ undistort($f_t$)
    \State Calculate feature velocity($un$ means undistort):
    \State \indent $F_t \leftarrow F_t + \frac{I_t(un(f_t)) - I_{t-1}(un(f_{t-1}))}{dt}$
    \State \Return $F_t$
\EndFunction
\end{algorithmic}
\end{algorithm}

\subsection{State Estimator}

In the State Estimator part, we use the sliding window-based optimization of VINS-Mono\cite{VINS}. We calculate the IMU measurement residual $r_B$ through IMU-preintegration and the visual measurement residual $r_C$ using the robust feature tracker within a sliding window, then compute the state $\mathcal{X}$ that minimizes these residuals\cite{VINS}.

\section{Experiments and Result}

\subsection{Experiment environment}

The experiments are performed using the EuRoC MAV dataset\cite{euroc} and the UMA Visual-Inertial dataset\cite{uma}.

The EuRoC MAV dataset\cite{euroc} includes various flight sequences collected in indoor environments and provides Ground Truth data. It is frequently used to evaluate the accuracy and robustness of VIO systems. Among these, we evaluate the hard sequences for each sequence type.

The UMA Visual-Inertial dataset\cite{uma} includes various sequences collected in both indoor and outdoor environments, allowing for performance evaluation under different lighting conditions and in textureless environments. It is useful for comprehensively evaluating the performance of VIO systems under diverse conditions. Since it does not provide Ground Truth, we conducted a qualitative evaluation.

Using both datasets, we compare and evaluate the performance of our method and VINS-Fusion(monocular-inertial odometry)\cite{VINS} using evo\cite{evo}.

\subsection{Experiment-1: Quantitative Evaluation of General Environments}

We evaluate three sequences from the EuRoC MAV dataset\cite{euroc}. Among these three sequences, V103 and V203 include aggressive motion and textureless environments.
\begin{itemize}
\item MH\_05\_difficult(MH05)
\item V1\_03\_difficult(V103)
\item V2\_03\_difficult(V203)
\end{itemize}

We compare the Ground Truth data with the Relative Pose Error (RPE) for evaluation. The experimental results are shown in Table 1. The performance across all three sequences is similar, indicating that our method achieves performance comparable to the existing system in general environments. Fig. 3 to Fig. 6 visualize the trajectory and RPE for each sequence.


\begin{table}[!t]
\setlength{\extrarowheight}{0.75ex}
\caption{Mean of RPE [m] in the EuRoC MAV datasets}
\label{euroc_res}
\centering
\begin{tabular}{|c|c|c|c|}
\hline
\diagbox[width=6.5pc,height=1.5pc]{~}{~} & MH05 & V103 & V203 \\
\hline
VINS-Fusion & \textbf{0.0045} & 0.0060 & 0.0088 \\
\hline
Ours & 0.0055 & \textbf{0.0049} & \textbf{0.0060} \\
\hline
\end{tabular}
\end{table}

\begin{figure}[!t]
\begin{center}
\label{traj_mh05}
\includegraphics[width=7.8cm]{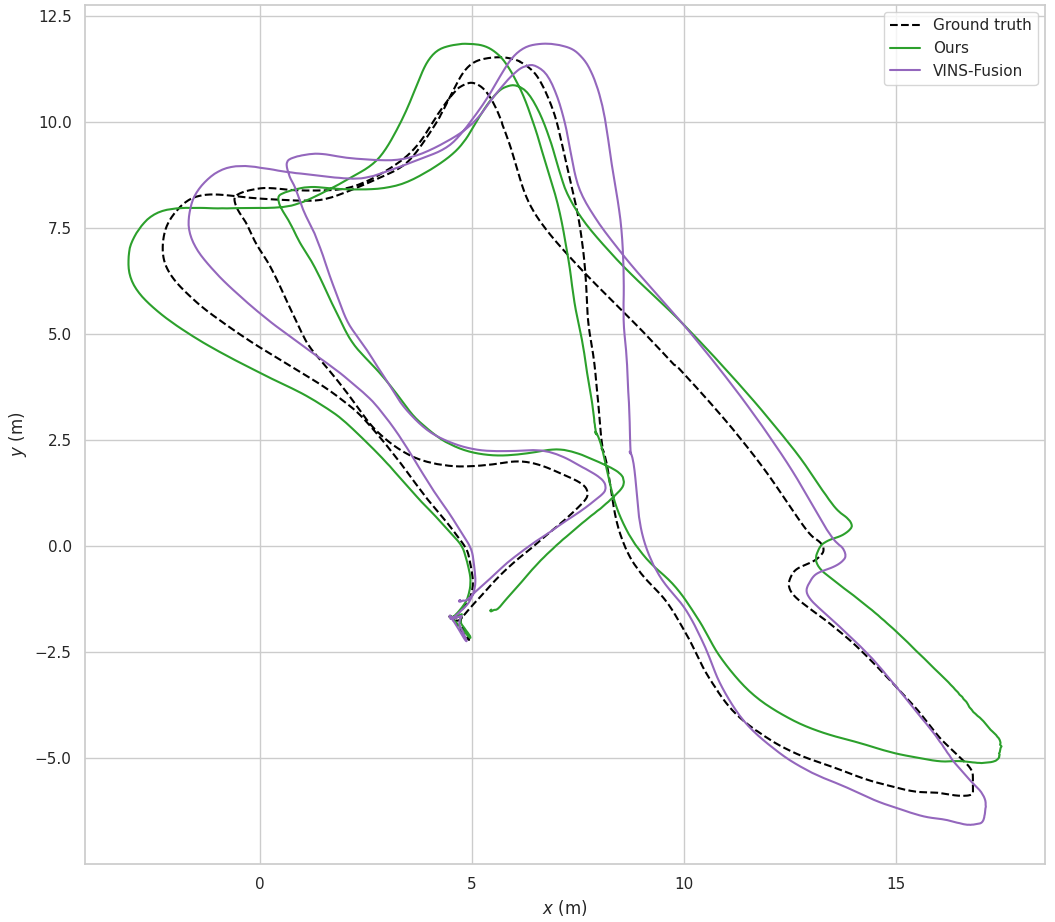}
\caption{Trajectory in MH05. Green line is result of ours, purple line is result of VINS-Fusion.}
\end{center}
\end{figure}

\begin{figure}[!t]
\begin{center}
\label{traj_v103}
\includegraphics[width=7.8cm]{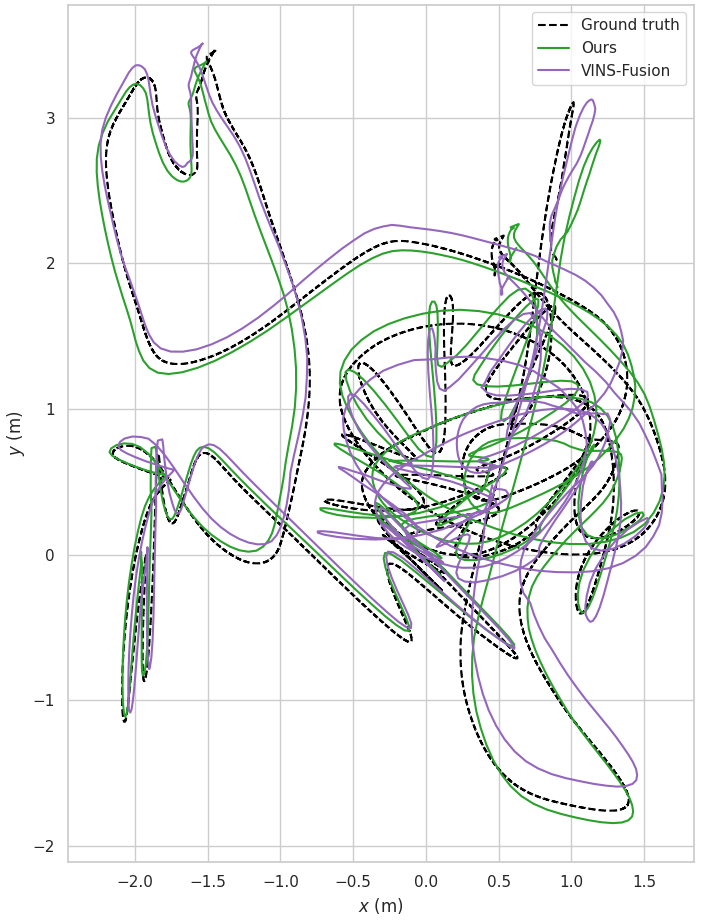}
\caption{Trajectory in V103. Green line is result of ours, purple line is result of VINS-Fusion.}
\end{center}
\end{figure}

\begin{figure}[!t]
\begin{center}
\label{traj_v203}
\includegraphics[width=7.8cm]{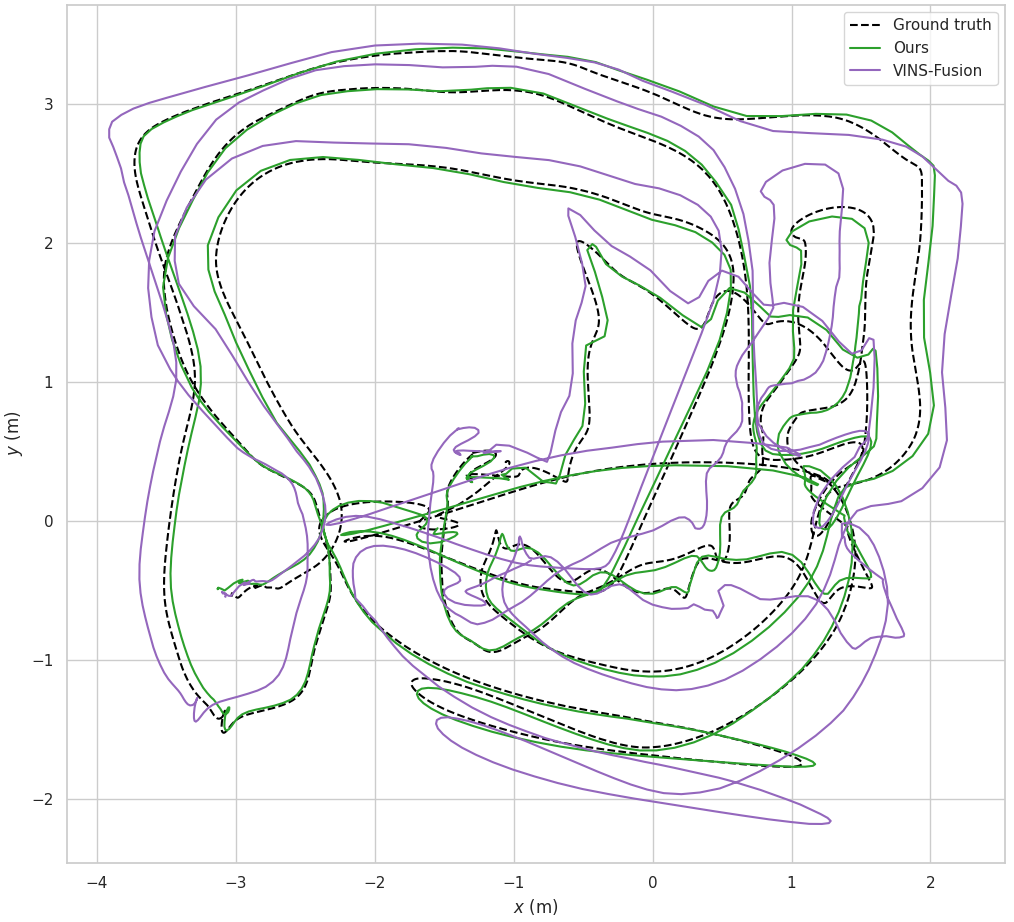}
\caption{Trajectory in V203. Green line is result of ours, purple line is result of VINS-Fusion.}
\end{center}
\end{figure}

\begin{figure}[!t]
\begin{center}
\includegraphics[width=2.5cm, height = 6cm]{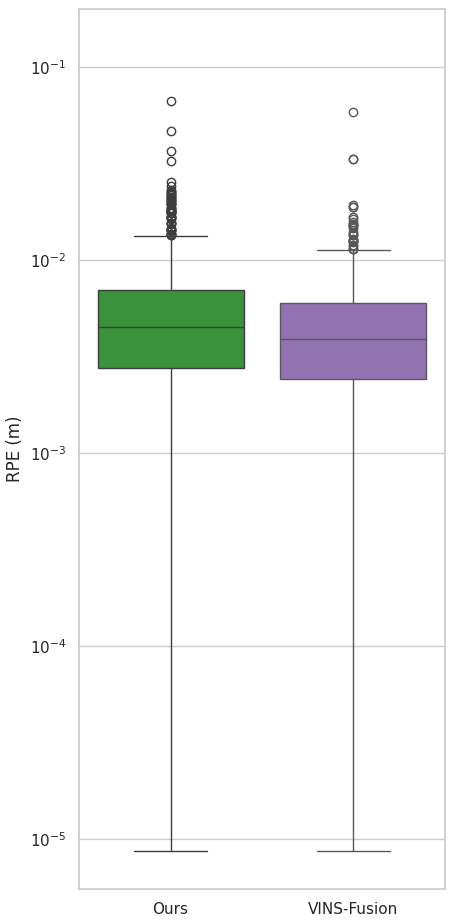}
\includegraphics[width=2.5cm, height = 6cm]{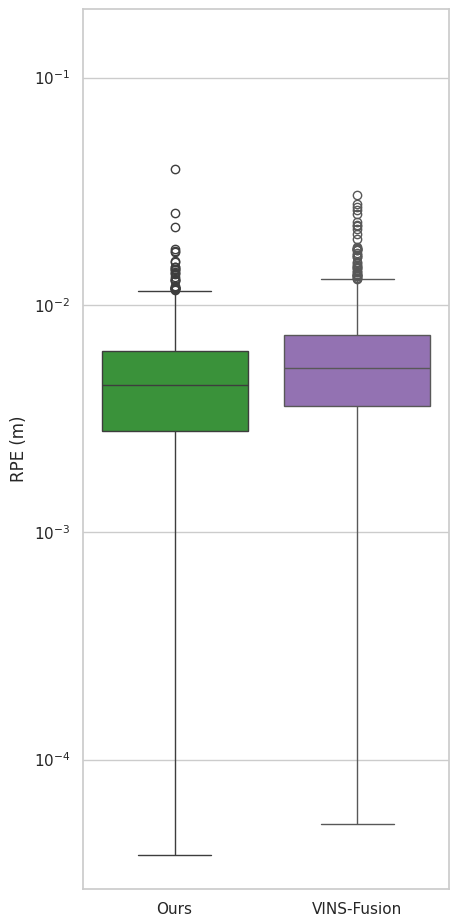}
\includegraphics[width=2.5cm, height = 6cm]{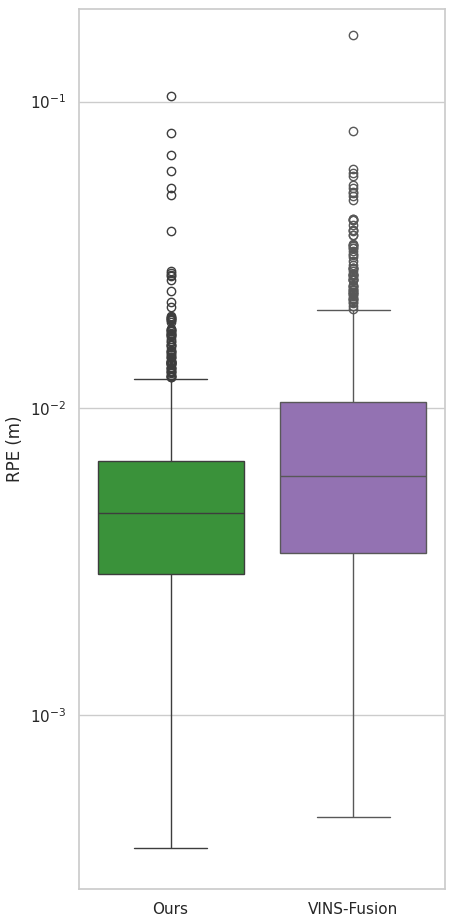}
\caption{RPE results: (Left) RPE in MH05. (Center) RPE in V103. (Right) RPE in V203. Green box is the result of ours, purple box is the result of VINS-Fusion.}
\label{fig:combined_rpe}
\end{center}
\end{figure}

\begin{figure}[!t]
\begin{center}
\label{uma_parking_ours}
\includegraphics[width=7.8cm]{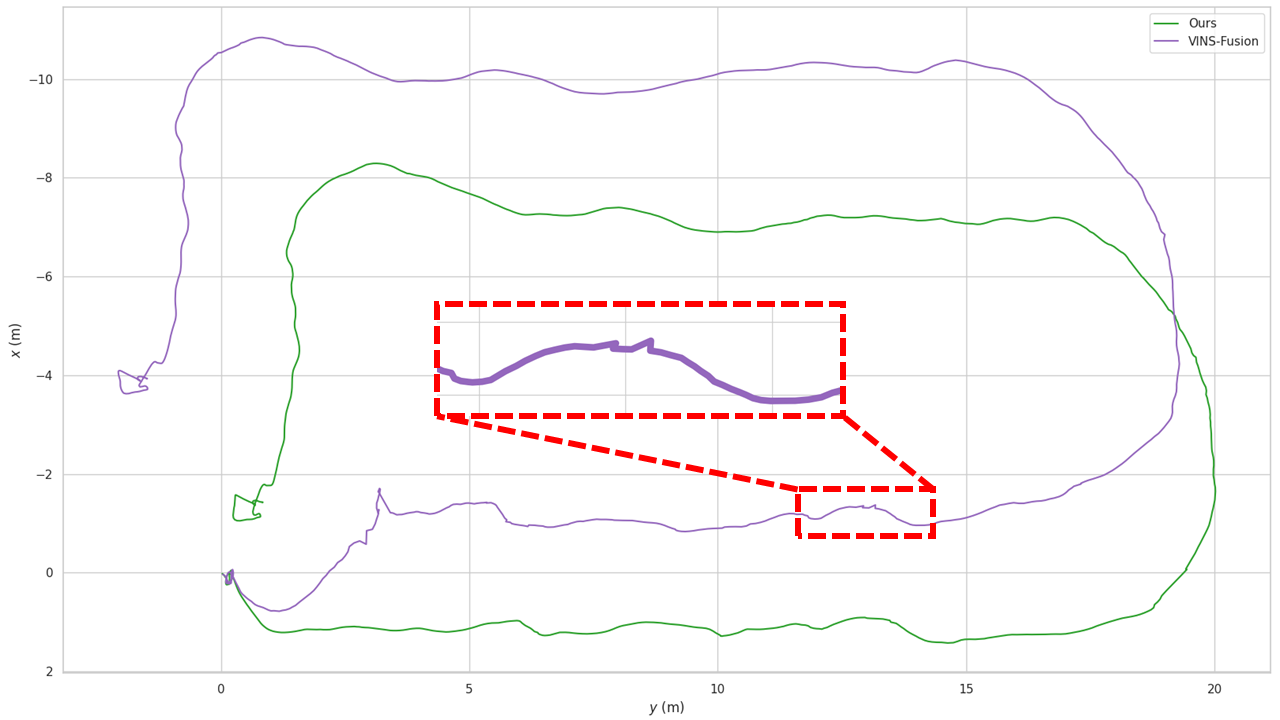}
\caption{Trajectory in UMA\_parking\_csc2. Green line is result of ours, purple line is result of VINS-Fusion.}
\end{center}
\end{figure}

\subsection{Experiment-2: Qualitative Evaluation of Challenging Environment}

We compared three sequences from the UMA Visual-Inertial dataset\cite{uma}. Each sequence has challenging environments, including dark settings, textureless areas, and dynamic illumination. All three sequences has same starting and ending points.

Fig. 7 shows the experimental results for UMA-parking\_csc2. Unlike our method, VINS-Fusion exhibits significant drift and noisy trajectory results. Additionally, our method demonstrates a trajectory closer to the origin, indicating better performance. For UMA\_parking\_eng1 (see Fig. 8), the sequence is affected by strong light, leading to unstable image sequences. VINS-Fusion shows divergence in the middle of the trajectory, whereas our method maintains a non-divergent trajectory. Lastly, for UMA\_fantasy\_csc1 (see Fig. 9), VINS-Fusion's trajectory contains a lot of noise, while our method remains more stable and closer to the origin.

As observed in the experiments, unstable and noisy trajectories are caused by the impact of incorrect feature tracking. Fig. 10 and Fig. 11 demonstrate that our feature tracking operates robustly in challenging environments.

\begin{figure}[!t]
\begin{center}
\label{uma_parking_ours}
\includegraphics[width=7.5cm]{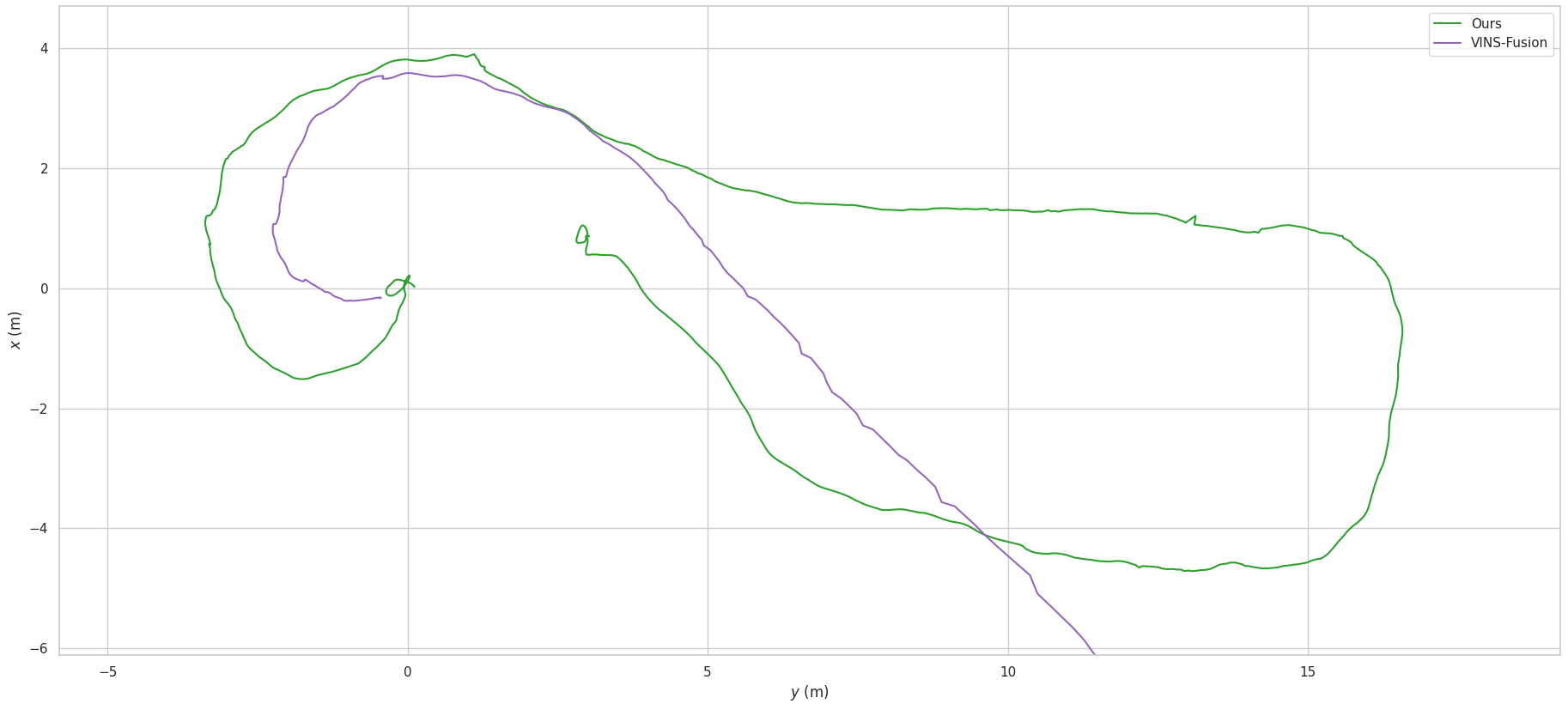}
\caption{Trajectory in UMA\_parking\_eng1. Green line is result of ours, purple line is result of VINS-Fusion.}
\end{center}
\end{figure}

\begin{figure}[!t]
\begin{center}
\label{uma_fantasy_ours}
\includegraphics[width=7.5cm]{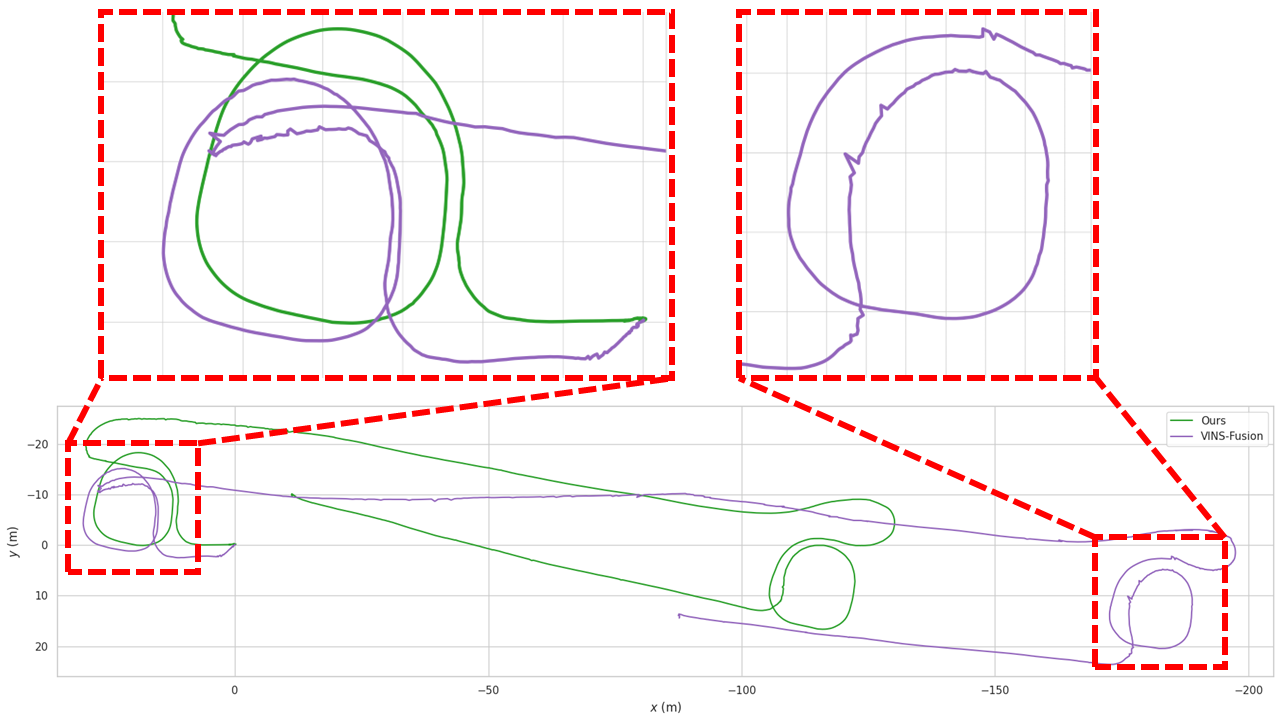}
\caption{Trajectory in UMA\_fantasy\_csc1. Green line is result of ours, purple line is result of VINS-Fusion.}
\end{center}
\end{figure}

\begin{figure}[!t]
\begin{center}
\label{ours_feat}
\includegraphics[width=7.0cm]{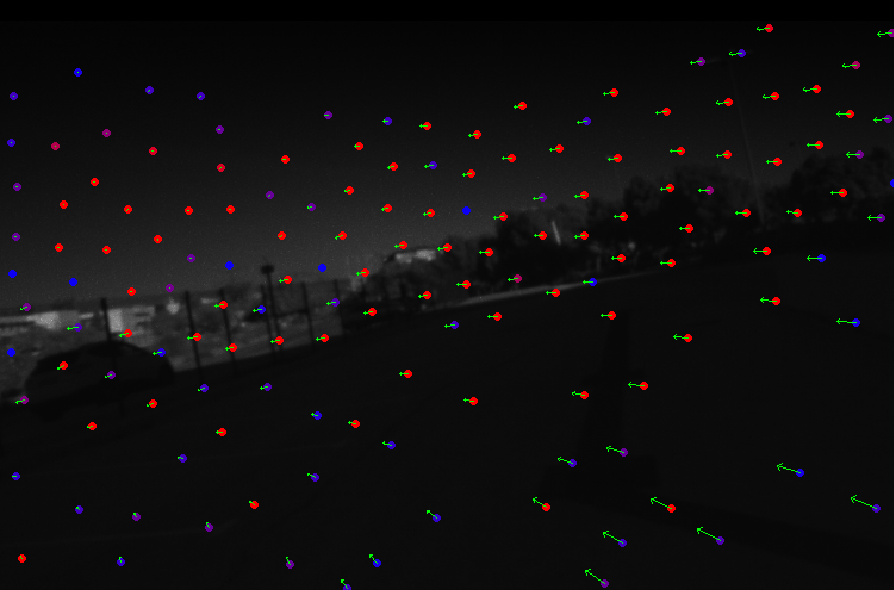}
\caption{Ours feature tracking image in challenging environment. As the feature points change from blue to red, it indicates that they have been tracked in more frames.}
\end{center}
\end{figure}

\begin{figure}[!t]
\begin{center}
\label{vins_feat}
\includegraphics[width=7.0cm]{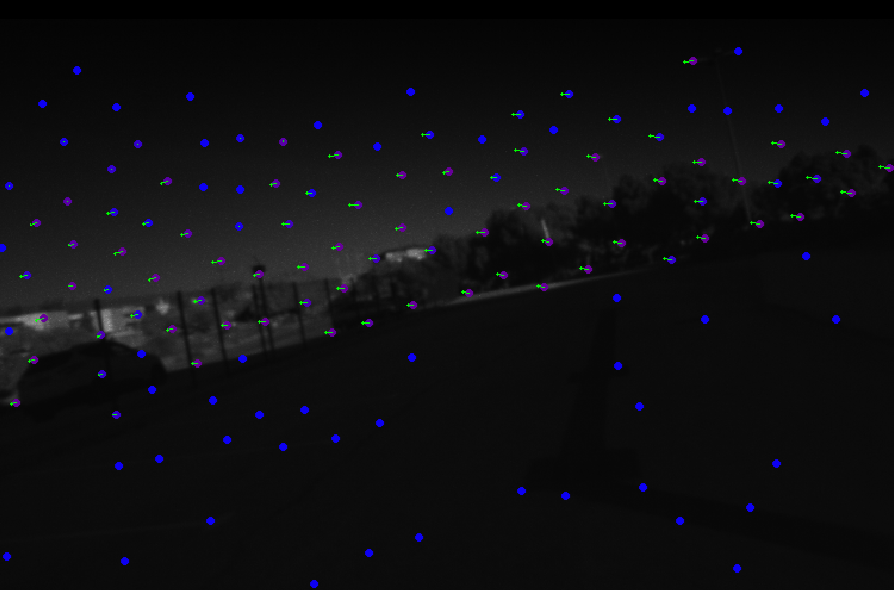}
\caption{VINS-Mono feature tracking image in challenging environment. As the feature points change from blue to red, it indicates that they have been tracked in more frames.}
\end{center}
\end{figure}

\section{Conclusion}

In this paper, we evaluated the odometry performance of our method and VINS-Mono\cite{VINS} using the EuRoC MAV dataset\cite{euroc} and the UMA Visual-Inertial dataset\cite{uma}. The experiments were conducted on challenging datasets to verify the robustness of the feature tracker, especially in environments with textureless and dynamic illumination. 

Our robust feature tracker demonstrated high feature tracking performance in textureless environments and situations with rapid changes in lighting. Additionally, it enabled stable odometry estimation with fewer outliers.

In conclusion, we have demonstrated that our method can significantly enhance the performance of the VINS-Mono\cite{VINS} system in challenging environment. By providing a more robust feature tracker, particularly in challenging situations such as environments with poor textures or rapid changes in lighting, this method shows great potential for applications in autonomous vehicles and robotics. Future research will focus on further optimizing computational costs and exploring the applicability of this method in various real-time applications.

\section*{Acknowledgement}

This research outputs are the part of the project “Disaster Field Investigation using Mobile Robot technology (Ⅱ)”, which is supported by the NDMI (National Disaster Management research Institute) under the project number NDMI-2024-06-02.

%
\bibliographystyle{unsrt}
\bibliography{ref}
%

\end{document}